\documentclass[accepted]{uai2023} 
\usepackage[american]{babel}

\usepackage{natbib} 
\usepackage{mathtools} 
\usepackage{booktabs} 
\usepackage{tikz} 
\usepackage{color}
\usepackage{algorithm}
\usepackage{algorithmic}
\usepackage{amsmath} 
\usepackage{amsthm}
\usepackage{amssymb}
\usepackage{graphicx}
\usepackage{epstopdf}
\usepackage{threeparttable}
\usepackage{multirow}
\usepackage{tabularx}

\newtheorem{mydef}{Definition}
\newtheorem{mytheo}{Theorem}

\newtheorem{myprop}{Proposition}

\usepackage[normalem]{ulem}
\useunder{\uline}{\ul}{}

\title{Scalable and Robust Tensor Ring Decomposition for Large-scale Data}

\author[1]{\href{yicong.he@ucf.edu}{Yicong He}{}}
\author[1,2]{\href{George.Atia@ucf.edu}{George K. Atia}}

\affil[1]{%
    Department of Electrical and Computer
    Engineering, University of Central Florida, Orlando, FL, 32816, USA.
}
\affil[2]{%
    Department of Computer
    Science, University of Central Florida, Orlando, FL, 32816, USA.
}
  
\begin{document}
\maketitle

\begin{abstract}
Tensor ring (TR) decomposition has recently received increased attention due to its superior expressive performance for high-order tensors. However, the applicability of traditional TR decomposition algorithms to real-world applications is hindered by prevalent large data sizes, missing entries, and corruption with outliers. In this work, we propose a scalable and robust TR decomposition algorithm capable of handling large-scale tensor data with missing entries and gross corruptions. We first develop a novel auto-weighted steepest descent method that can adaptively fill the missing entries and identify the outliers during the decomposition process. Further, taking advantage of the tensor ring model, we develop a novel fast Gram matrix computation (FGMC) approach and a randomized subtensor sketching (RStS) strategy which yield significant reduction in storage and computational complexity. Experimental results demonstrate that the proposed method outperforms existing TR decomposition methods in the presence of outliers, and runs significantly faster than existing robust tensor completion algorithms.
\end{abstract}

\section{Introduction}
The demand for multi-dimensional data processing has led to increased attention in multi-way data analysis using tensor representations. Tensors generalize matrices in higher dimensions and can be expressed in a compressed form using a sequence of operations on simpler tensors through tensor decomposition \citep{kolda2009tensor,sidiropoulos2017tensor}. Tensor decomposition, an extension of matrix factorization \citep{koren2009matrix} to higher dimensions, plays a vital role in tensor analysis, as many real-world data, such as videos and MRI images, contain latent and redundant structures \citep{chen2013simultaneous,li2017online}. Various tensor decomposition models, such as Tucker \citep{tucker1966some}, CANDECOMP/PARAFAC (CP) \citep{carroll1970analysis,yamaguchi2017tensor}, tensor train (TT) \citep{oseledets2011tensor}, and tensor ring (TR) \citep{zhao2016tensor}, have been proposed by developing different tensor latent spaces. The tensor ring decomposition, the primary focus of this work, offers notable perceived advantages such as high compression performance for high-order tensors, enhanced performance in completion tasks with high missing rates \citep{wang2017efficient,yu2020low}, and more generality and flexibility than the TT model.

Despite its recognized advantages, TR decomposition faces challenges that limit its usefulness in real-world applications. One such challenge is scalability, as traditional TR decomposition methods become computationally and storage-intensive as tensor size increases. To address this limitation, various algorithms have been developed to improve efficiency and scalability, including randomized methods \citep{malik2021sampling,yuan2019randomized,ahmadi2020randomized}.

Another challenge is robustness to missing entries and outliers, which requires a robust tensor completion approach. Existing TR decomposition methods \citep{zhao2016tensor,malik2021sampling,yuan2019randomized,ahmadi2020randomized} based on second-order error residuals perform poorly in the presence of outliers. While more robust norms, such as the $\ell_1$-norm, are commonly used in machine learning, the non-smoothness and non-differentiability of the $\ell_1$-norm at zero make it difficult to realize scalable versions of existing algorithms.



The primary focus of this work is to simultaneously address the two foregoing challenges by developing a scalable and robust TR decomposition algorithm. We first develop a new full-scale robust TR decomposition method. An information theoretic learning-based similarity measure called correntropy \citep{liu2007correntropy} is introduced to the TR decomposition problem and a new differentiable correntropy-based cost function is proposed. Utilizing a half-quadratic technique \citep{nikolova2005analysis}, the non-convex problem is reformulated as an auto-weighted decomposition problem that adaptively alleviates the effect of outliers. To solve the problem, we introduce a scaled steepest descent method \citep{tanner2016low}, that lends itself to further acceleration through a scalable scheme. By exploiting the structure of the TR model, we develop two acceleration methods for the proposed robust approach, namely, fast Gram matrix computation (FGMC) and randomized subtensor sketching (RStS). Utilizing FGMC reduces the complexity of Gram matrix computation from exponential to linear complexity in the order of the tensor. With RStS, only a small sketch of data is used per iteration, which makes the algorithm scalable to large tensor data. The main contributions of the paper are summarized as follows: 

1) We develop a new scalable and robust TR decomposition method. Using correntropy error measure and leveraging an HQ technique, an efficient auto-weighted robust TR decomposition (AWRTRD) algorithm is proposed. 

2) By developing a novel fast Gram matrix computation (FGMC) method and a randomized subtensor sketching (RStS) strategy, we develop a more scalable version of AWRTRD, which significantly reduces both the computational time and storage requirements.


3) We conduct experiments on image and video data, verifying the robustness of our proposed algorithms compared with existing TR decomposition algorithms. Moreover, we perform experiments on completion tasks that demonstrate that our proposed algorithm can handle large-scale tensor completion with significantly less time and memory cost than existing robust tensor completion algorithms.

\section{related work}

\textbf{Scalable TR decomposition:}
Scalable TR decomposition methods are necessary for solving large-scale decomposition problems. In \citet{malik2021sampling}, a sampling-based TR alternating least-squares (TRALS-S) method was proposed using leverage scores to accelerate the ALS procedure in TRALS \citep{zhao2016tensor}. In \citet{yuan2019randomized}, a randomized projection-based TRALS (rTRALS) method was proposed, which uses random projection on every mode of the tensor. In \citet{ahmadi2020randomized}, a series of fast TR decomposition algorithms based on randomized singular value decomposition (SVD) were developed. Although these methods have demonstrated desired performance in large-scale TR decomposition, they are unable to handle cases where some entries are missing or perturbed by outliers.

\textbf{Robust tensor completion:} To mitigate the impact of outliers, various robust tensor completion algorithms have been developed under different tensor decomposition models \citep{jiang2019robust,huang2020robust,goldfarb2014robust,yang2015robust}. For the TR decomposition model, \citet{huang2020robust} proposed a robust $\ell_1$-regularized tensor ring nuclear norm ($\ell_1$-TRNN) completion algorithm, where the Frobenius norm of the error measure in TRNN \citep{yu2019tensor} is replaced with the robust $\ell_1$-norm. Additionally, a $\ell_{p,\epsilon}$-regularized tensor ring completion ($\ell_{p,\epsilon}$-TRC) algorithm was developed \citep{li2021robust}. These robust tensor completion algorithms, however, cannot be easily extended to scalable versions as they utilize non-differentiable $\ell_1$ or $\ell_{p,\epsilon}$ norm, and optimize the nuclear norm on unfolding matrices of the tensor.

\section{Preliminaries}

\noindent\textbf{Notation.} Uppercase script letters are used to denote tensors (e.g., ${\mathcal{X}}$), and boldface letters to denote matrices (e.g., ${\mathbf{X}}$). An $N$-order tensor is defined as $\mathcal{X}\in{\mathbb{R}}^{I_1\times \cdots \times I_N}$, where $I_i,i\in[N]:=\{1,\ldots,N\}$, is the dimension of the $i$-th way of the tensor, and $\mathcal{X}_{i_1\ldots i_N}$ denotes the $(i_1,i_2,\ldots,i_N)$-th entry of tensor $\mathcal{X}$. For a $3$-rd order tensor (i.e., $N=3$), the notation ${{\cal{X}}}(:,:,i),{{\cal{X}}}(:,i,:),{{\cal{X}}}(i,:,:)$ denotes the frontal, lateral, and horizontal slices of $\mathbf{\cal{X}}$, respectively. The Frobenius norm of tensor ${\cal X}$ is defined as $\|\mathcal{X}\|_F=\sqrt{\sum_{i_1\ldots i_N}|{{\mathcal{X}}_{i_1\ldots i_N}}|^2}$. $\operatorname{Tr}(\cdot)$ is the matrix trace operator. Next, we provide a brief overview of the definition of TR decomposition and some results that will be utilized in this paper.
\begin{mydef}
[TR Decomposition \citep{zhao2016tensor}] Given TR rank $[r_1,\ldots,r_N]$, in TR decomposition, a high-order tensor $\mathcal{X} \in \mathbb{R}^{I_{1} \times \cdots \times I_{N}}$ is represented as a sequence of circularly contracted 3-order core tensors $\mathcal{Z}_k\in\mathbb{R}^{r_k\times I_k\times r_{k+1}}, k=1,\ldots,N$, with $r_{N+1}=r_1$. Specifically, the element-wise relation of tensor $\mathcal{X}$ and its TR core tensors $\{\mathcal{Z}_k\}_{k=1}^N$ is defined as
\[
\mathcal{X}_{i_{1} \ldots i_{N}}=\operatorname{Tr}\left(\prod_{k=1}^N\mathbf{Z}_{k}\left(i_{k}\right)\right)\:,
\]
where $\mathbf{Z}_{k}\left(i_{k}\right):=\mathcal{Z}_k(:,i_k,:)$ denotes the $i_{k}$-th lateral slice matrix of the latent tensor $\mathcal{Z}_{k}$, 
which is of size $r_{k} \times r_{k+1}$.
\label{def:TRdecomp}
\end{mydef}

\begin{mydef}
[Tensor core merging \citep{zhao2016tensor}]
Let $\mathcal{X}=\Re\left(\mathcal{Z}_{1}, \mathcal{Z}_{2}, \ldots, \mathcal{Z}_{N}\right)$ be a TR representation of an $N$-order tensor, where $\mathcal{Z}_{k} \in \mathbb{R}^{r_{k} \times I_{k} \times r_{k+1}}, k=1, \ldots, N$, is a sequence of cores. Since the adjacent cores $\mathcal{Z}_{k}$ and $\mathcal{Z}_{k+1}$ have an equivalent mode size $r_{k+1}$, they can be merged into a single core by multilinear products, which is defined by $\mathcal{Z}^{(k, k+1)} \in \mathbb{R}^{r_{k} \times I_{k} I_{k+1} \times r_{k+2}}$ whose lateral slice matrices are given by
$$
\mathbf{Z}^{(k, k+1)}\left(\textcolor{black}{\overline{i_{k} i_{k+1}}}\right)=\mathbf{Z}_{k}\left(i_{k}\right) \mathbf{Z}_{k+1}\left(i_{k+1}\right)
$$
\textcolor{black}{where $\overline{i_{1} i_{2}\ldots i_{N}}=i_{1}+\left(i_2-1\right)I_1+\cdots+\left(i_{N}-1\right) I_{1} I_{2} \ldots I_{N-1}$.}
\end{mydef}

\begin{mytheo}
[\citep{zhao2016tensor}]
Given a TR decomposition of tensor $\mathcal{X}=\Re\left(\mathcal{Z}_{1}, \ldots, \mathcal{Z}_{N}\right)$, its mode-$k$ unfolding matrix $\mathbf{X}_{[k]}$ can be written as
\[
\mathbf{X}_{[k]}=\textcolor{black}{\mathbf{Z}_{k(2)}}(\mathbf{Z}_{[2]}^{\neq k})^T\:,
\]
where $\mathcal{Z}^{\neq k}\in\mathbb{R}^{r_{k+1} \times \prod_{1, j \neq k}^{N} I_{j} \times r_{k}}$ is a subchain obtained by merging all cores except $\mathcal{Z}^{k}$, whose lateral slice matrices are defined by 
$$
\mathbf{Z}^{\neq k}\left(\overline{i_{k+1} \cdots i_{N} i_{1} \ldots i_{k-1}}\right)\!=\!\prod_{k+1}^{N} \!\mathbf{Z}_{j}\left(i_{j}\right) \prod_{1}^{k-1} \mathbf{Z}_{j}\left(i_{j}\right)\:.
$$
The mode-$n$ unfolding of $\mathcal{X}$ is the matrix $\mathbf{X}_{[n]} \in$ $\mathbb{R}^{I_{n} \times \prod_{j \neq n} I_{j}}$ defined element-wise via
$$
\mathbf{X}_{[n]}\left(i_{n}, \overline{i_{n+1} \cdots i_{N} i_{1} \cdots i_{n-1}}\right) \stackrel{\text { def }}{=} \mathcal{X}\left(i_{1}, \ldots, i_{N}\right)\:,
$$
and $\mathbf{X}_{(n)}$ is the classical mode-$n$ unfolding of $\mathcal{X}$, that is, the matrix $\mathbf{X}_{(n)} \in \mathbb{R}^{I_{n} \times \prod_{j \neq n} I_{j}}$ defined element-wise via
$$
\mathbf{X}_{(n)}\left(i_{n}, \overline{i_{1} \cdots i_{n-1} i_{n+1} \cdots i_{N}}\right) \stackrel{\text { def }}{=} \mathcal{X}\left(i_{1}, \ldots, i_{N}\right)\:.
$$
\label{thm1}
\end{mytheo}

\section{Proposed Approach}
\subsection{Correntropy-based TR Decomposition}

As shown in Definition \ref{def:TRdecomp}, TR decomposition amounts to finding a set of core tensors $\{\mathcal{Z}_k\}_{k=1}^N$ that can approximate $\mathcal{X}$ given the values of the TR-rank $[r_1,\ldots,r_N]$. In practice, the optimization problem can be formulated as
\begin{equation}
\min _{\mathcal{Z}_{1}, \ldots, \mathcal{Z}_{N}}\left\|{\mathcal{X}}-\Re(\mathcal{Z}_{1}, \ldots, \mathcal{Z}_{N})\right\|_{F}^2\:.
\label{eq:TRALS}
\end{equation}
A common method for solving \eqref{eq:TRALS} is the tensor ring-based alternating least-squares (TRALS) \citep{zhao2016tensor}. When $\mathcal{X}$ is partially observed (i.e., there are missing entries in $\mathcal{X}$), the objective function is extended as 
\begin{equation}
\min _{\mathcal{Z}_{1}, \ldots, \mathcal{Z}_{N}}\left\|\mathcal{P}\circ({\mathcal{X}}-\Re(\mathcal{Z}_{1}, \ldots, \mathcal{Z}_{N}))\right\|_{F}^2
\label{eq:TRWOPT}
\end{equation}
where $\mathcal{P}\in\{0,1\}^{I_1\times\cdots\times I_N}$ is a binary mask tensor that indicates the locations of the observed entries of $\mathcal{X}$ (entries corresponding to the observed entries in $\mathcal{X}$ are set to $1$, and the others are set to $0$). 

In addition to missing entries, real-world data is often corrupted with outliers, which can result in unreliable observations. This has motivated further research on robust tensor decomposition and completion methods. The predominant measure of error in robust tensor decomposition/completion is the $\ell_1$-norm of the error residual \citep{gu2014robust,huang2020robust,wang2020robust}. 
However, the non-smoothness and non-differentiability of the $\ell_1$-norm at zero presents a challenge in extending this formulation to scalable methods for large-scale data.

To enhance robustness and scalability, we formulate a new robust TR decomposition optimization problem using the correntropy measure. Correntropy \citep{liu2007correntropy} is a local and nonlinear similarity measure defined by the kernel width $\sigma$. Given two $N$-dimensional discrete vectors $\mathbf{x}$ and $\mathbf{y}$, the correntropy is measured as
\begin{equation}
V (\mathbf{x}, \mathbf{y})= \frac{1}{N} \kappa_{\sigma}(x_i-y_i)\:,
\end{equation}
\par
where $\kappa_{\sigma}$ is a kernel function with kernel width $\sigma$. It has been demonstrated that with a proper choice of the kernel function and kernel width, correntropy-based error measure is less sensitive to outliers compared to the $\ell_2$-norm \citep{liu2007correntropy,wang2016correntropy}.

In this work, we introduce the correntropy measure in TR decomposition. 
Specifically, we replace the second-order error in \eqref{eq:TRWOPT} with correntropy at the element-wise level and use the Gaussian kernel as the kernel function. This leads to the following new optimization problem based on correntropy:
\begin{equation}
\max _{\mathcal{Z}_{1}, \ldots, \mathcal{Z}_{N}}G_{\sigma}(\mathcal{P}\circ({\mathcal{X}}-\Re(\mathcal{Z}_{1}, \ldots, \mathcal{Z}_{N})))
\label{eq:corr}
\end{equation}
where $G_{\sigma}(\mathcal{X})=\sum_{i_1\ldots i_N}g_{\sigma}(\mathcal{X}_{i_1\ldots i_N})$ and $g_{\sigma}(x)=\sigma^2\exp(-\frac{x^2}{2\sigma^2})$. Note that the objective function is maximized since the correntropy becomes large when the error residual is small. Next, we develop a new algorithm to solve the problem in \eqref{eq:corr} efficiently, while also leaving room for further modifications to enhance its scalability.

\subsection{Auto-weighted scaled steepest descent method for robust TR decomposition}

To efficiently solve \eqref{eq:corr}, we leverage a half-quadratic (HQ) technique which has been applied to non-quadratic optimization in previous works \citep{he2011robust,he2014robust}. In particular, according to Proposition 1 in \citet{he2011robust}, there exists a convex
conjugated function $\varphi$ of $g_{\sigma}(x)$ such that 
\begin{equation}
\begin{aligned}
\max_x g_{\sigma}(x)=\min_{x,w} wx^2+\varphi(w)\:,\\
\end{aligned}
\label{eq:hq_w2}
\end{equation}
where the optimal solution of $w$ in the right hand side (RHS) of \eqref{eq:hq_w2} is given by ${w}^*=\frac{g'_{\sigma}(x)}{x}$.
Thus, maximizing $g_\sigma (x)$ in terms of $x$ is equivalent to minimizing an augmented cost function in an enlarged parameter space $\{x,w\}$. By substituting \eqref{eq:hq_w2} in \eqref{eq:corr}, the complex optimization problem can be solved using the following optimization problem
\begin{equation}
\begin{aligned}
\min_{\mathcal{Z}_1,\ldots,\mathcal{Z}_N,\mathcal{W}}\hspace{-.5mm}\frac{1}{2}\|\sqrt{\mathcal{W}}\hspace{-.5mm}\circ\hspace{-.5mm}\mathcal{P}\hspace{-.5mm}\circ\hspace{-.5mm}(\mathcal{X}\hspace{-.5mm}-\hspace{-.5mm}\Re(\mathcal{Z}_{1}, \ldots, \mathcal{Z}_{N}))\|_F^2\hspace{-.5mm}+\hspace{-.5mm}\Psi(\mathcal{W}),
\end{aligned}
\label{eq:tc_hq}
\end{equation}
where $\Psi \left( {{\cal{W}}}\right)=\sum\nolimits_{i_1\ldots i_N}\varphi \left( {{{\cal{W}}_{i_1\ldots i_N}}} \right)$. Problem \eqref{eq:tc_hq} can be regarded as an auto-weighted TR decomposition. Specifically, the weighting tensor ${\mathcal{W}}$ automatically assigns different weights to each entry based on the error residual. According to the property of the Gaussian function, given a proper kernel width $\sigma$, a large error residual caused by an outlier may result in a small weight, thereby alleviating the impact of outliers. Further, when $\sigma\rightarrow\infty$, $\frac{g'_{\sigma}(x)}{x}$ approaches $1$, thus all the entries of $\mathcal{W}$ become $1$. In this case, the optimization problem in \eqref{eq:tc_hq} reduces to the traditional TR decomposition problem in \eqref{eq:TRWOPT}. 

Next, we propose a new scaled steepest descent method to solve \eqref{eq:tc_hq}. The solution process is summarized next.

1) Updating $\mathcal{W}$: according to \eqref{eq:hq_w2}, each element $\mathcal{W}_{i_1\ldots i_N}$ corresponding to its observed entry can be obtained as
\begin{equation}
\begin{aligned}
\mathcal{W}_{i_1\ldots i_N}&=\frac{g'_{\sigma}(\mathcal{E}_{i_1\ldots i_N})}{\mathcal{E}_{i_1\ldots i_N}},\\
\text{for}~(i_1\ldots i_N)&\in\{(i_1\ldots i_N)|\mathcal{P}_{i_1\ldots i_N}\!=\!1\}
\end{aligned}
\label{eq:updatew}
\end{equation}
where $\mathcal{E}=\mathcal{X}-\Re(\mathcal{Z}_{1}, \ldots, \mathcal{Z}_{N})$. It should be noted that updating $\mathcal{W}_{i_1\ldots i_N}$ for unobserved entries does not affect the results due to multiplication with $\mathcal{P}$ in \eqref{eq:tc_hq}. Therefore, in the following part we update all entries of $\mathcal{W}$.

2) Updating $\{\mathcal{Z}_k\}_{k=1}^N$: According to Theorem \ref{thm1} and \eqref{eq:tc_hq}, for $\mathcal{Z}_k$ with any $k\in[1,N]$, by fixing $\mathcal{W}$ and $\{\mathcal{Z}_j\}_{j=1,j\neq k}^N$, $\mathcal{Z}_k$ can be obtained using the following minimization problem
\begin{equation}
\begin{aligned}
\min_{\mathbf{Z}_{k(2)}}\left\|\sqrt{\mathbf{W}}_{[k]}\circ\mathbf{P}_{[k]}\circ(\mathbf{X}_{[k]}\!-\!\mathbf{Z}_{k(2)}(\mathbf{Z}_{[2]}^{\neq k})^{T})\right\|_{F}^2\:.
\end{aligned}
\label{eq:TRALSz}
\end{equation}
It is difficult to obtain a closed-form solution to \eqref{eq:TRALSz} due to the existence of $\mathcal{P}$. Instead, we apply the gradient descent method. To this end, taking the derivative w.r.t. $\mathbf{Z}_{k(2)}$, we obtain the gradient in terms of $\mathbf{Z}_{k(2)}$ as
\begin{equation}
d(\mathbf{Z}_{k(2)})=\left(\mathbf{W}_{[k]}\circ\mathbf{P}_{[k]}\circ\left(\mathbf{X}_{[k]}-\mathbf{Z}_{k(2)}(\mathbf{Z}_{[2]}^{\neq k})^{T}\right)\right)\mathbf{Z}_{[2]}^{\neq k}\:.
\label{eq:sd}
\end{equation}
In general, the above gradient descent method can be directly applied by cyclically updating the core tensors $\mathcal{Z}_{k}$ using $d(\mathbf{Z}_{k(2)})$. However, the convergence rate of the gradient descent method could be slow in practice. To improve convergence, we introduce a scaled steepest descent method \citep{tanner2016low} to solve \eqref{eq:TRALSz}. In particular, the scaled gradient in terms of $\mathbf{Z}_{k(2)}$ is
\begin{equation}
\begin{aligned}
h(\mathbf{Z}_{k(2)})=d(\mathbf{Z}_{k(2)})\left((\mathbf{Z}_{[2]}^{\neq k})^{T}\mathbf{Z}_{[2]}^{\neq k}+\lambda\mathbf{I}\right)^{-1}\:.
\end{aligned}
\label{eq:sg}
\end{equation}
The regularization parameter $\lambda$ is utilized to avoid singularity and can be set to a sufficiently small value. Finally, $\mathcal{Z}_{k}$ is updated as
\begin{equation}
\mathcal{Z}_{k}=\mathcal{Z}_{k}-\eta_k\operatorname{fold}(h(\mathbf{Z}_{k(2)}))\:,
\label{eq:updatessd}
\end{equation}
where the operator $\operatorname{fold(.)}$ tensorizes its matrix argument, and the step-size $\eta_k$ is set using exact line-search as
\begin{equation}
\begin{aligned}
\eta_k=\frac{\langle d(\mathbf{Z}_{k(2)}),h(\mathbf{Z}_{k(2)})\rangle}{\left\|\sqrt{\mathbf{W}}_{[k]}\circ\mathbf{P}_{[k]}\circ \left(h(\mathbf{Z}_{k(2)})(\mathbf{Z}_{[2]}^{\neq k})^{T}\right)\right\|_F^2}\:,
\end{aligned}
\end{equation}
where $\langle\mathbf{A},\mathbf{B}\rangle$ is the inner product of matrices $\mathbf{A}$ and $\mathbf{B}$. 

We name the proposed method auto-weighted robust tensor ring decomposition (AWRTRD). Next, we develop two novel scalable strategies to accelerate the computation of AWRTRD.

\subsection{Scalable strategies for AWRTRD}

Although AWRTRD enhances robustness and mitigates the effect of outliers, several challenges  limit its applicability to large-scale data. In particular, the matrices $\mathbf{X}_{[k]}$ and $\mathbf{Z}_{[2]}^{\neq k}$ are of size ${I_{k} \times \prod_{j \neq k} I_{j}}$ and ${r_{k}r_{k+1} \times \prod_{1, j \neq k}^{N} I_{j} }$, respectively. When the tensor size is large, the matrices $\mathbf{X}_{[k]}$ and $\mathbf{Z}_{[2]}^{\neq k}$ become very large, and the calculations in \eqref{eq:sd} and \eqref{eq:sg} can become computational bottlenecks, making it difficult to use AWRTRD with large-scale data. 
More specifically, one needs to compute two large-scale matrix multiplications: 1) $\mathbf{Y}\mathbf{Z}_{[2]}^{\neq k}$ in \eqref{eq:sd} with $\mathbf{Y}=\mathbf{W}_{[k]}\circ\mathbf{P}_{[k]}\circ\left(\mathbf{X}_{[k]}-\mathbf{Z}_{k(2)}(\mathbf{Z}_{[2]}^{\neq k})^{T}\right)$; 2) $(\mathbf{Z}_{[2]}^{\neq k})^{T}\mathbf{Z}_{[2]}^{\neq k}$ in \eqref{eq:sg}. To date, no prior work has specifically addressed the acceleration of these operations. Therefore, in this section, leveraging the TR model structure along with randomized subtensor sketching, we devise two novel strategies to accelerate the above two computation operations.

\subsubsection{Fast Gram matrix computation (FGMC)}

In this section, we develop a fast Gram matrix computation (FGMC) of $\mathbf{Z}_{[2]}^{\neq k,T}\mathbf{Z}_{[2]}^{\neq k}$,  by exploiting the structure of the TR model. For simplicity, in the following we use $\mathbf{G}_{\mathcal{Z}}$ to denote $\mathbf{Z}_{[2]}^{T}\mathbf{Z}_{[2]}$. According to tensor core merging of two core tensors $\mathcal{Z}_{k}$ and $\mathcal{Z}_{k+1}$ in Definition 2, we establish the following result. The proof is provided as supplementary material. 

\begin{myprop}
\label{prop_2}
Let $\mathcal{Z}_{k} \in$ $\mathbb{R}^{r_{k} \times I_{k} \times r_{k+1}}, k=1, \ldots, N$, be $3$-rd order tensors. Defining $\mathcal{Z}^{\leq c}\in\mathbb{R}^{r_1\times\prod_{k=1}^c I_k\times r_{c+1}}$ as a subchain obtained by merging $c$ cores $\{\mathcal{Z}_{k}\}_{k=1}^c$, i.e.,
$$
\mathbf{Z}^{\leq c}\left(\overline{i_{1} \cdots i_{c}}\right)=\prod_{k=1}^{c} \mathbf{Z}_{k}\left(i_{k}\right)\:,
$$
the Gram matrix of $\mathbf{Z}_{[2]}^{\leq c}$ can be computed as
\begin{equation}
\begin{aligned}
\mathbf{G}_{\mathcal{Z}^{\leq c}}=\mathbf{Z}_{[2]}^{\leq c,T}\mathbf{Z}_{[2]}^{\leq c}=\Phi\left(\prod_{k=1}^{c}\mathbf{Q}_{k}\right)\:,
\end{aligned}
\label{eq:FGMC}
\end{equation}
where $\mathbf{Q}_k(:,i\times r_{k+1}+j)=\operatorname{vec}\{\left(\mathcal{Z}_{k}(:,:,i)\right)\mathcal{Z}_{k}(:,:,j)^T\}$ for $k>1$ and
$$ \mathbf{Q}_1(:,i\times r_2+j)\!=\!\left\{
\begin{aligned}
\operatorname{vec}\{\left(\mathcal{Z}_{1}(:,:,i)\right)\!\mathcal{Z}_{1}(:,:,j)^T\},&~c~\text{is even} \\
\operatorname{vec}\{\left(\mathcal{Z}_{1}(:,:,j)\right)\!\mathcal{Z}_{1}(:,:,i)^T\},&~c~\text{is odd}
\end{aligned}
\right.
$$
\end{myprop}

Proposition \ref{prop_2} allows us to compute $\mathbf{G}_{\mathcal{Z}^{\neq k}}$ without explicitly calculating $\mathbf{Z}_{[2]}^{\neq k}$. Further, $\mathbf{Q}_k$ is only related to $\mathcal{Z}_k$, hence 
can be updated independently. FGMC yields considerable computation and storage gains; for the simple case where $r_1\!=\!\cdots\!=\!r_N\!=\!r$ and $I_1\!=\!\cdots\!=\!I_N\!=\!I$, traditional computation of $\mathbf{G}_{\mathcal{Z}^{\neq k}}$ requires $\mathcal{O}(I^Nr^4+r^6)$ in time complexity and $\mathcal{O}(I^{N}r^2)$ in storage complexity, while the proposed FGMC method requires only $\mathcal{O}(NIr^4+r^6)$ in time complexity and $\mathcal{O}(Ir^4)$ in storage complexity, which is linear in the tensor order $N$.

\subsubsection{Randomized subtensor sketching (RStS)} 

Unlike $\mathbf{G}_{\mathcal{Z}^{\neq k}}$ whose complexity can be reduced by taking advantage of the TR model, directly accelerating the computation  of \eqref{eq:sd} is challenging. 
In \citet{malik2021sampling}, a leverage score sampling-based strategy is applied to TRALS. In each ALS iteration, the leverage score is computed for each row of $\mathbf{Z}_{[2]}^{\neq k}$. Then, the rows are sampled with probability proportional to the leverage scores. The computation of $\mathbf{X}_{[k]}\mathbf{Z}_{[2]}^{\neq k}$ is reduced to $(\mathbf{X}_{[k]})_{\mathcal{I}}(\mathbf{Z}_{[2]}^{\neq k})_{\mathcal{I}}$, where $\mathcal{I}$ is the index set of the sampled rows. Although this method reduces the data used per iteration compared to the traditional TRALS algorithm, the computation of the leverage scores is costly as it requires computing an SVD per iteration. 


Inspired by the random sampling method \citep{vervliet2015randomized} for CP decomposition, we introduce a randomized subtensor sketching strategy for TR decomposition. 
Specifically, given an $N$th-order tensor $\mathcal{X}\in\mathbb{R}^{I_1\times I_2\times\cdots\times I_N}$, defining $\mathcal{I}$ as the sample index set, and $\mathcal{I}_k$ as the sample index set of the $k$-th tensor dimension, we sample the tensor along each dimension according to $\mathcal{I}_k,k=1,\ldots,N$, and obtain the sampled subtensor $\mathcal{X}_{\mathcal{I}}\in\mathbb{R}^{s_1\times\ldots\times s_N}$, where $s_k=|\mathcal{I}_k|$ is the sample size for the $k$-th order. It is not hard to conclude that
\begin{equation}
\mathcal{X}_{\mathcal{I}}=\Re\left((\mathcal{Z}_1)_{\mathcal{I}_1},(\mathcal{Z}_2)_{\mathcal{I}_2},\ldots,(\mathcal{Z}_N)_{\mathcal{I}_N}\right)\:,
\end{equation}
where $(\mathcal{Z}_k)_{\mathcal{I}_k}\in\mathbb{R}^{r_k\times s_k \times r_{k+1}}$ is a sampled core subtensor of $\mathcal{Z}_{k}$ obtained by sampling lateral slices of $\mathcal{Z}_{k}$ with index set $\mathcal{I}_k$. 
Compared with directly sampling rows from $\mathbf{Z}_{[2]}^{\neq k}$, the proposed method restricts the sampling on a tensor sketch $\mathcal{X}_{\mathcal{I}}$ and intentionally requires enough sampling along all dimensions, and also requires fewer core tensors to construct $\mathcal{X}_{\mathcal{I}}$ for the same sample size. The superior performance of the proposed sampling method is verified in the experimental results.

\subsection{Scalable AWRTRD using FGMC and RStS}

Given the FGMC and RStS methods described above, we can readily describe our scalable version of the proposed AWRTRD algorithm. In particular, at each iteration, core tensors are cyclically updated from $\mathcal{Z}_1$ to $\mathcal{Z}_N$. Specifically, by leveraging RStS, the gradient $d(\mathbf{Z}_{k(2)})$ can be approximated by the gradient using a subtensor $\mathcal{X}_\mathcal{I}$ sampled from the original tensor $\mathcal{X}$. For $k\in[1,N]$, given a sample parameter $J$, we set $s_k=I_k$ and $s_j= \lceil J^{\frac{1}{N-1}}/\prod_{i\neq k,j}I_i\rceil$ for $j\in[1,\ldots,k-1,k+1,\ldots,N]$. Then, for $k=1,\ldots,N$, we randomly and uniformly select $s_k$ lateral slices of $\mathcal{Z}_k$ to get the sampled core subtensors $\{(\mathcal{Z}_k)_{\mathcal{I}_k}\}_{k=1}^N$ and corresponding sampled subtensor $\mathcal{X}_\mathcal{I}$. 

The optimization is similar to AWRTRD. First, we compute the corresponding sampled entries of the weight tensor $\mathcal{W}$ using \eqref{eq:updatew}, and get a new sampled weight tensor $\mathcal{W}_\mathcal{I}$. Then, the gradient is computed as 
\begin{equation}
\begin{aligned}
\!d_{\mathcal{I}}(\mathbf{Z}_{k(2)})=&\left(\mathbf{W}_{\mathcal{I}[k]}\circ\mathbf{P}_{\mathcal{I}[k]}\circ\left(\mathbf{X}_{\mathcal{I}[k]}\right.\right.\\
&\left.\left.-\mathbf{Z}_{k(2)}(\mathbf{Z}_{\mathcal{I}[2]}^{\neq k})^T\right)\right)\mathbf{Z}_{\mathcal{I}[2]}^{\neq k}
\end{aligned}
\label{eq:hatd}
\end{equation}
where $\mathcal{Z}_{\mathcal{I}}^{\neq k}\in\mathbb{R}^{r_{k+1} \times \prod_{1, j \neq k}^{N} s_{j} \times r_{k}}$ is a subchain obtained by merging all sampled cores except $(\mathcal{Z}_k)_{\mathcal{I}_k}$. Subsequently, the scaled gradient is computed by
\begin{equation}
\begin{aligned}
h_{\mathcal{I}}(\mathbf{Z}_{k(2)})&=d_{\mathcal{I}}(\mathbf{Z}_{k(2)})\left((\mathbf{Z}_{[2]}^{\neq k})^T\mathbf{Z}_{[2]}^{\neq k}+\lambda\mathbf{I}\right)^{-1}\:.
\label{eq:hI}
\end{aligned}
\end{equation}
Note that the term $\mathbf{G}_{\mathcal{Z}^{\neq k}}=\mathbf{Z}_{[2]}^{\neq k,T}\mathbf{Z}_{[2]}^{\neq k}$ is still computed using the full core tensors, such that the global information can be preserved. As described in Proposition 2, $\mathbf{G}_{\mathcal{Z}^{\neq k}}$ can be efficiently computed using FGMC. Finally, $\mathcal{Z}_{k}$ is updated as
\begin{equation}
\mathcal{Z}_{k}=\mathcal{Z}_{k}-\eta_k\operatorname{fold}(h_{\mathcal{I}}(\mathbf{Z}_{k(2)}))\:,
\label{eq:ftrbcn}
\end{equation}
with the step-size set using exact line-search as
\begin{equation}
\begin{aligned}
\eta_k=\frac{\langle d_{\mathcal{I}}(\mathbf{Z}_{k(2)}),h_{\mathcal{I}}(\mathbf{Z}_{k(2)})\rangle}{\left\|\sqrt{\mathbf{W}}_{\mathcal{I}[k]}\circ\mathbf{P}_{\mathcal{I}[k]}\circ \left(h_{\mathcal{I}}(\mathbf{Z}_{k(2)})(\mathbf{Z}_{\mathcal{I}[2]}^{\neq k})^T\right)\right\|_F^2}\:.
\end{aligned}
\end{equation}


We term the above algorithm Scalable AWRTRD (SAWRTRD), and its pseudocode is presented in Algorithm 1. 

\begin{algorithm}
\caption{Scalable AWRTRD (SAWRTRD)}
\begin{algorithmic}[1]
\REQUIRE Tensor $\mathcal{X}\in\mathbb{R}^{I_1\times I_2\times\cdots\times I_N}$, TR ranks $\{r_k\}_{k=1}^N$, sample parameter $J$, kernel width $\sigma$, maximum iteration number $C$, error tolerance $\epsilon$.
\STATE Initialize $\mathcal{Z}_{k} \in \mathbb{R}^{r_{k} \times I_{k} \times r_{k+1}} $ and corresponding $\mathbf{Q}_{k}, k=1, \ldots, N $, $t=0$. Set $\mathcal{W}$ as the all-one tensor.
\REPEAT
\FOR {$k=1,\ldots,N$}
\STATE Randomly generate index set $\mathcal{I}_i\subseteq\{1,...,I_k\}$ for $i\neq k$ with size $s_i= \lceil J^{\frac{1}{N-1}}/\prod_{j\neq i,k}I_j \rceil$, set $\mathcal{I}_k=\{1,\ldots,I_k\}$.
\STATE Obtain subtensor $\mathcal{X}_{\mathcal{I}}$, $\mathcal{W}_{\mathcal{I}}$ and $\{(\mathcal{Z}_k)_{\mathcal{I}_k}\}_{i=1}^N$.
\STATE Update corresponding entries of $\mathcal{W}_{\mathcal{I}}$ using \eqref{eq:updatew}.
\STATE Compute ${d}_{\mathcal{I}}(\mathbf{Z}_{k(2)})$ using \eqref{eq:hatd}.
\STATE Obtain $\mathbf{G}_{\mathcal{Z}^{\neq k}}$ with $\{\mathbf{Q}_{i}\}_{i=1,i\neq k}^N$ using \eqref{eq:FGMC}.
\STATE Update $\mathcal{Z}_{k}$ using \eqref{eq:ftrbcn}.
\STATE Update $\mathbf{Q}_{k}$ according to $\mathcal{Z}_{k}$.
\ENDFOR
\STATE Compute matrix $\!\mathbf{D}^t\!=\!\mathbf{Z}_{\!N(2)\!}\mathbf{G}_{\mathcal{Z}^{\neq N}}\mathbf{Z}^T_{N(2)}$
\STATE Compute relative error $e=\|\mathbf{D}^t-\mathbf{D}^{t-1}\|_F/\|\mathbf{D}^t\|_F$
\STATE $t=t+1$
\UNTIL $t=C$ or $e<\epsilon$.
\ENSURE TR cores $\mathcal{Z}_{k}, k=1, \ldots, N$.
\end{algorithmic}
\label{alg:ftrbcnscalable}
\end{algorithm}

\section{Computational complexity}
\label{sec:complexity}

For simplicity, we assume the TR rank $r_1\!=\!\cdots\!=\!r_N\!=\!r$, and the data size $I_1\!=\!\cdots\!=\!I_N\!=\!I$. For AWRTRD, the computations of $d(\mathbf{Z}_{k(2)})$ and $\mathbf{G}_{\mathcal{Z}^{\neq k}}$ have complexity $\mathcal{O}(I^Nr^2)$ and $\mathcal{O}(I^Nr^4+r^6)$, respectively. Therefore, the complexity of AWRTRD is $\mathcal{O}(NI^Nr^4+Nr^6)$. For SAWRTRD with sample parameter $J$, the computation of $\hat{d}(\mathbf{Z}_{k(2)})$ has complexity $\mathcal{O}(NIJr^2)$. Hence, combining the FMGC analyzed in the previous section, the complexity of SAWRTRD is $\mathcal{O}(NIJr^2+N^2Ir^4+Nr^6)$. The time complexities of different TR decomposition algorithms are shown in Table \ref{tab:complexity}. We assume the projection dimensions of rTRALS are all $K$. As shown, the complexity of SAWRTRD is smaller than AWRTRD. Further, TRALS-S performs SVD on unfolding matrices $\mathbf{Z}_{k(2)},k=1,\ldots,N$ at each iteration, thus becomes less efficient as $I$ increases. 

\begin{table}[htbp]
\caption{Time complexity of different TR decomposition algorithms}
\label{tab:complexity}
\centering
{\begin{tabular}{ccc}
\toprule
Algorithms & time complexity \\ \midrule
TRSVD      & $\mathcal{O}(I^{N+1}+I^Nr^3)$    \\
TRALS      & $\mathcal{O}(NI^Nr^4+Nr^6)$    \\
TRSVD-R      & $\mathcal{O}(I^Nr^2)$    \\
rTRALS     & $\mathcal{O}(NK^Nr^4+Nr^6)$   \\
TRALS-S    & $\mathcal{O}(NIJr^4+Nr^6)$   \\
AWRTRD     & $\mathcal{O}(NI^Nr^4+Nr^6)$     \\
SAWRTRD   & $\mathcal{O}(NIJr^2+N^2Ir^4+Nr^6)$    \\ \bottomrule
\end{tabular}%
}
\end{table}

\section{Experimental Results}
In this section, we present experimental results to demonstrate the performance of the proposed methods. The evaluation includes two tasks: robust TR decomposition (no missing entries) and robust tensor completion. For TR decomposition, we compare the performance to the traditional TR decomposition methods TRALS and TRSVD \citep{zhao2016tensor}, and three scalable TR decomposition methods rTRALS \citep{yuan2019randomized}, TRALS-S\footnote{https://github.com/OsmanMalik/TRALS-sampled} \citep{malik2021sampling} and TRSVD-R \citep{ahmadi2020randomized}. For robust tensor completion, we compare the performance with $\ell_1$-regularized sum of nuclear norm ($\ell_1$-SNN)\footnote{https://tonyzqin.wordpress.com/research} \citep{goldfarb2014robust}, $\ell_1$-regularized tensor nuclear norm ($\ell_1$-TNN) \citep{jiang2019robust}, $\ell_1$ regularized tensor ring nuclear norm ($\ell_1$-TRNN)\footnote{https://github.com/HuyanHuang/Robust-Low-rank-Tensor-Ring-Completion} \citep{huang2020robust}, $\ell_{p,\epsilon}$-regularized tensor ring completion ($\ell_{p,\epsilon}$-TRC) \citep{li2021robust} and transformed nuclear norm-based total variation (TNTV)\footnote{https://github.com/xjzhang008/TNTV} \citep{qiu2021robust}.

\begin{figure}[tb]
\centering
\includegraphics[width=\linewidth]{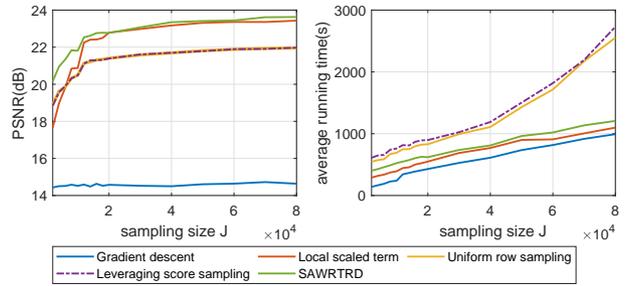}
\caption{Average PSNR and running times versus sampling parameter $J$ using different strategies.}
\label{fig:ablation}
\end{figure}

Both the decomposition and completion performance are evaluated using the Peak Signal-to-Noise Ratio (PSNR) between the original data and the recovered data. For each experiment, the PSNR value is averaged over 20 Monte Carlo runs with different noise realizations and missing locations. For the kernel width $\sigma$ in the proposed methods, we apply the adaptive kernel width selection strategy described in \citet{he2019robust}. The maximum number of iterations for all algorithms is set to $30$. The error tolerance $\epsilon$ of all algorithms is set to $10^{-3}$. All other parameters of the algorithms that we compare to are set to achieve their best performance in our experiments. All experiments were performed using MATLAB R2021a on a desktop PC with a 2.5-GHz processor and 32GB of RAM.

\subsection{Ablation experiments}

In this part, we carry out ablation experiments to verify the feasibility and advantage of the proposed strategies and also analyze the performance under different sampling sizes. The experiment is carried out using a color video `flamingo' with resolution $1920\times 1080$ chosen from DAVIS 2016 dataset\footnote{https://davischallenge.org/davis2016/code.html}. The first $50$ frames are selected, so the video data can be represented as a $4$-dimensional tensor with size $1920\times1080\times3\times50$. The data values are rescaled to $[0,1]$, and $30\%$ of the pixels are randomly and uniformly selected as the observed pixels. Then salt and pepper noise is added to the observed entries with probability $0.2$. 

To demonstrate the advantage of the proposed methods, we develop four additional algorithms using different strategies. The first algorithm, referred to as `gradient descent', uses the traditional gradient $d_{\mathcal{I}}(\mathbf{Z}_{k(2)})$ in \eqref{eq:hatd} instead of $h_{\mathcal{I}}(\mathbf{Z}_{k(2)})$ in \eqref{eq:hI}. The second algorithm, termed `local scaled term', applies the local scaled term $\mathbf{G}_{\mathcal{Z}_{\mathcal{I}}^{\neq k}}$ from sampled core tensors $\{(\mathcal{Z}_k)_{\mathcal{I}_k}\}_{k=1}^N$ to \eqref{eq:hI} instead of the global scaled term $\mathbf{G}_{\mathcal{Z}}^{\neq k}$. For the third algorithm, termed `uniform row sampling', $\mathbf{Z}_{\mathcal{I}[2]}^{\neq k}$ is obtained using uniform row sampling \citep{malik2021sampling} instead of RStS. In the fourth algorithm, referred to as `leverage score sampling', the leverage score row sampling used in TRALS-S is directly applied. It should also be noticed that SAWRTRD without using FMGC is not included in the comparison, as it will be out of memory. Fig. \ref{fig:ablation} depicts the average PSNR and running times for five algorithms under different sampling sizes. The TR rank for all methods is set to [80, 80, 3, 20]. As can be seen, the values of PSNR for all algorithms are relatively stable for sample parameter $J\geq4\times10^4$. Also, the proposed RStS sampling strategy yields higher PSNR than the row sampling methods as well as lower computational cost. Further, the global scaled term outperforms the local scaled term, especially for small sampling size.

\begin{figure}[tb]
\centering
\includegraphics[width=\linewidth]{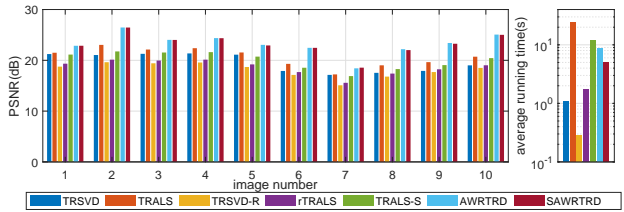}
\caption{PSNR (left) for each image using different TR decomposition algorithms. Right: Average running time (right) of each algorithm on 10 images.}
\label{fig:image_decomp_UAI}
\end{figure}

\begin{figure}[tb]
\centering
\includegraphics[width=\linewidth]{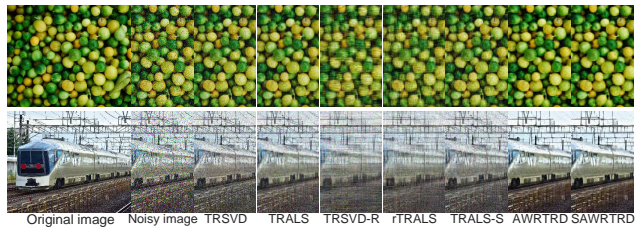}
\caption{Recovered (cropped) images by different TR decomposition algorithms. Top: Image 2. Bottom: Image 7.}
\label{fig:image_decomp_example}
\end{figure}

\begin{figure}[tb]
\centering
\includegraphics[width=\linewidth]{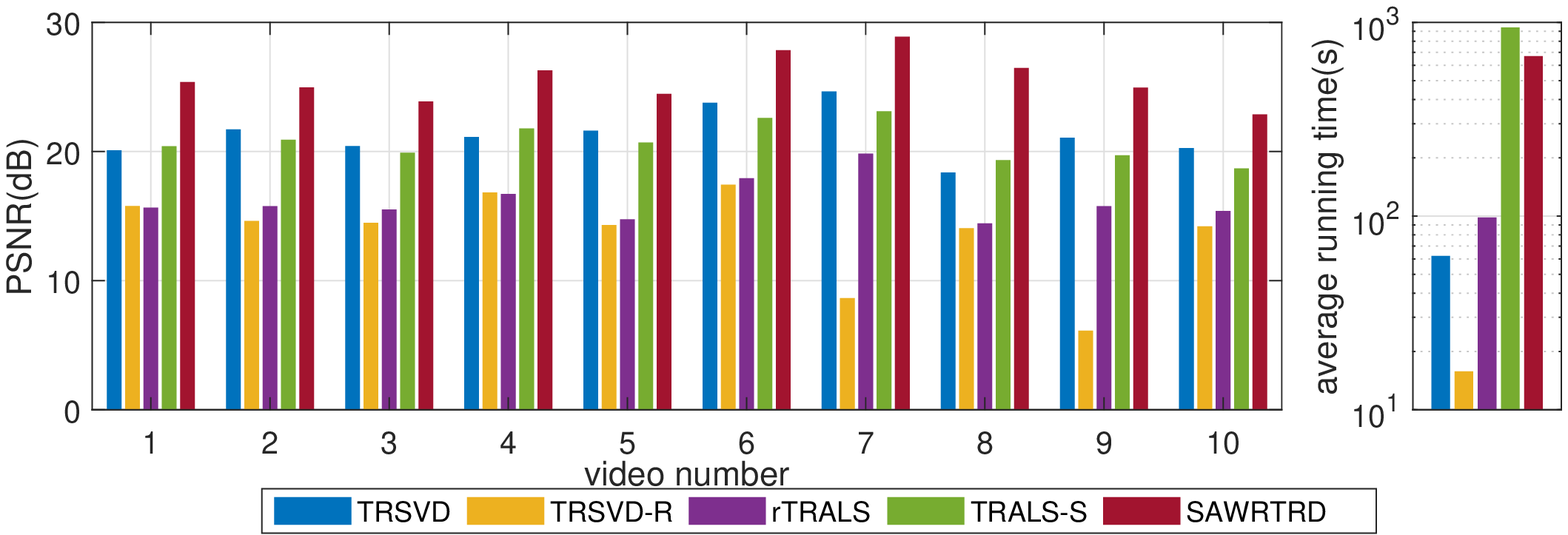}
\caption{PSNR (left) for each video using different TR decomposition algorithms. Right: Average running time (right) of each algorithm on 10 videos.}
\label{fig:video_decomp_UAI}
\end{figure}

\begin{figure}[tb]
\centering
\includegraphics[width=\linewidth]{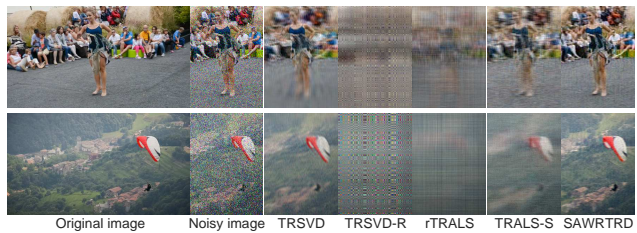}
\caption{Recovered $10^{th}$ (cropped) frames by different TR decomposition algorithms. Top: Video 2. Bottom: Video 7.}
\label{fig:video_decomp_example}
\end{figure}

\subsection{High-resolution image/video TR decomposition with noise}

In this part, we investigate the performance of TR decomposition in the presence of outliers. We first compare the decomposition performance on color image data. Specifically, 10 images are randomly selected from the DIV2K dataset\footnote{https://data.vision.ee.ethz.ch/cvl/DIV2K} \citep{Agustsson_2017_CVPR_Workshops}. Two representative images are shown in Fig. \ref{fig:image_decomp_example}. The height and width of the images are about $1350$ pixels and $2000$, respectively. 

The values of the image data are rescaled to $[0,1]$, then two types of noise are added to the 10 images. Specifically, for the first 5 images, the noise is generated from a two-component Gaussian mixture model (GMM) with probability density function (PDF) $0.8N(0,10^{-3})+0.2N(0,0.5)$, where the latter term denotes the occurrence of outliers. For the last 5 images, $20\%$ of the pixels are perturbed with salt and pepper noise. The sampling parameter $J$ for SAWRTRD is set to $3\times10^4$. The TR rank is set to $[20,20,3]$. For TRALS-S, the number of sampling rows for each mode is set to the same as SAWRTRD. The PSNR and average running time for different TR decomposition algorithms are shown in Fig. \ref{fig:image_decomp_UAI}. As shown, the PSNR values of AWRTRD and SAWRTRD are always higher than the other algorithms. An example of the recovered images from the obtained core tensors is shown in Fig. \ref{fig:image_decomp_example}. As shown, the images recovered by AWRTRD and SAWRTRD are visually better than the other algorithms.  

\begin{figure}[tb]
\centering
\includegraphics[width=\linewidth]{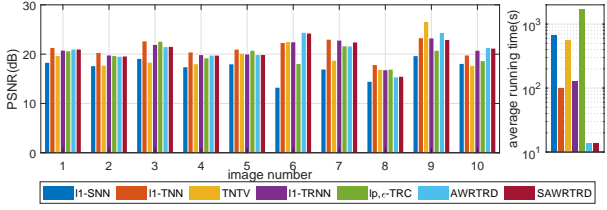}
\caption{PSNR (left) for each image using different robust tensor completion algorithms. Right: Average running time (right) of each algorithm on 10 images.}
\label{fig:image_completion_UAI}
\end{figure}

\begin{figure}[tb]
\centering
\includegraphics[width=\linewidth]{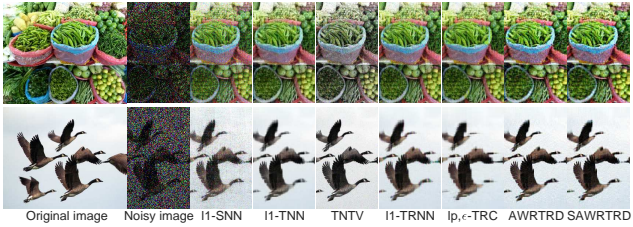}
\caption{Recovered (cropped) images by different tensor completion algorithms. Top: Image 2. Bottom: Image 9.}
\label{fig:image_completion_example}
\end{figure}

\begin{figure}[tb]
\centering
\includegraphics[width=\linewidth]{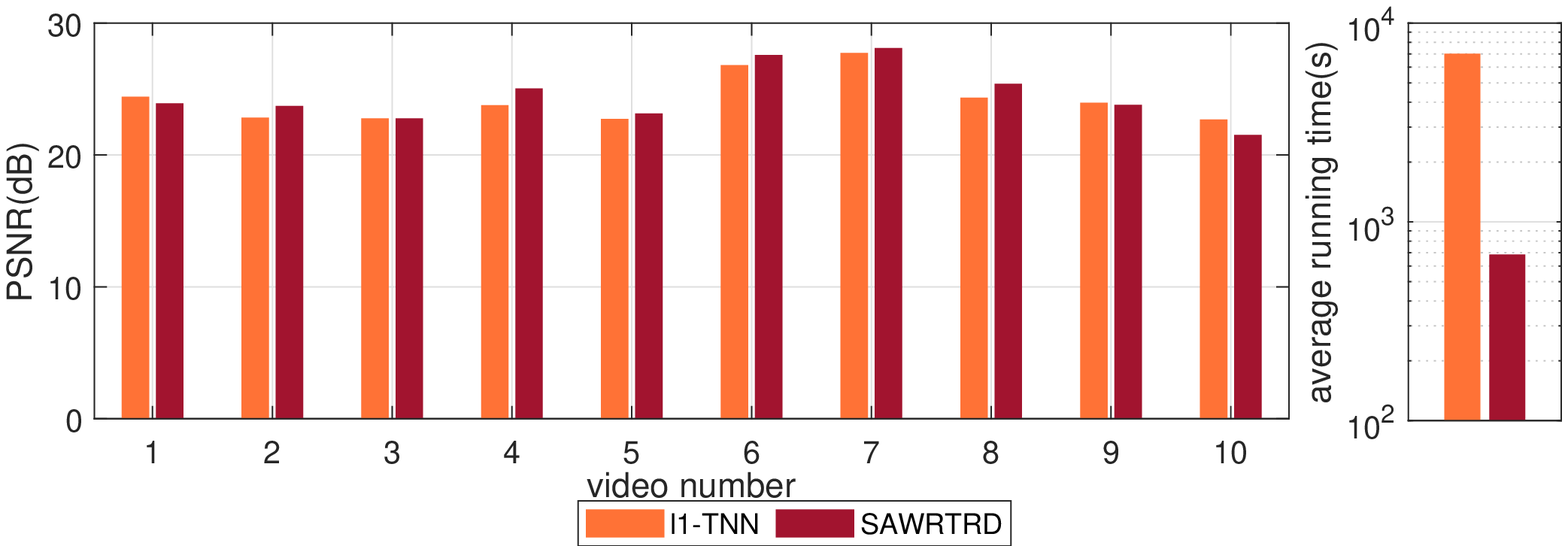}
\caption{PSNR (left) for each video using different robust tensor completion algorithms. Right: Average running time (right) of each algorithm on 10 videos.}
\label{fig:video_completion_UAI}
\end{figure}

\begin{figure}[tb]
\centering
\includegraphics[width=\linewidth]{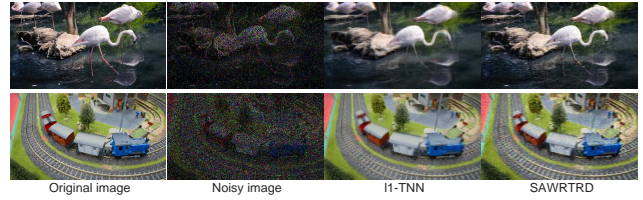}
\caption{Recovered $10^{th}$ (cropped) frames by different tensor completion algorithms. Top: Video 5. Bottom: Video 10.}
\label{fig:video_completion_example}
\end{figure}

Then, we compare the decomposition performance of the proposed methods on video data. 10 videos with resolution $1920\times 1080$ are randomly chosen from the DAVIS 2016 dataset and the first $50$ frames are selected to form the tensor data. Representative frames from two videos are shown in Fig. \ref{fig:video_decomp_example}. The values of the video data are rescaled to $[0,1]$, and the noises are added with distribution the same as in the previous experiment. Fig. \ref{fig:video_decomp_UAI} shows the PSNR and average running times using different robust tensor completion algorithms. Note that TRALS and AWRTRD are out of memory in the experiment, such that their results are not shown. As can be seen, the proposed SAWRTRD achieves significantly higher PSNR than other algorithms for all videos. Further, Fig. \ref{fig:video_decomp_example} illustrates the $10^{th}$ recovered frame of two videos. It can be seen that TRSVD-R and rTRALS fail to recover the frames, and the frames recovered by SAWRTRD have the clearest textures. 

\subsection{Image/video completion with noise}

In this part, we compare the tensor completion performance of different robust tensor completion algorithms. Similar to the previous experiment, we randomly select 10 color images from DIV2K dataset, and 10 videos with the first $50$ frames from DAVIS 2016 dataset. The noises are added with distributions that are the same as in the previous experiment. Further, for each image and video, $30\%$ of the pixels are randomly and uniformly selected as the observed pixels. The sample parameter $J$ and the TR rank are set to $3\times 10^4$ and [80, 80, 3, 20], respectively. Fig.\ref{fig:image_completion_UAI} and Fig.\ref{fig:video_completion_UAI} report the PSNR and running times using different algorithms on images and videos, respectively. We should remark that only SAWRTRD and $\ell_1$-TNN can handle the video completion task while other algorithms run out of memory. As can be seen, in image completion, the proposed method can achieve comparable robust completion performance to other algorithms with significantly lower computational costs. For video completion, the performance of SAWRTRD and $\ell_1$-TNN are comparable, while the time cost of SAWRTRD is significantly smaller than $\ell_1$-TNN. Examples of the completed images and video frames are also given in Fig. \ref{fig:image_completion_example} and Fig.\ref{fig:video_completion_example}, respectively. Finally, we should remark that, unlike other robust tensor completion methods, our algorithm can also get compact TR representations of the images and videos. 

\section{Conclusion}
We proposed a scalable and robust approach to TR decomposition. 
By introducing correntropy as the error measure, the proposed method can alleviate the impact of large outliers. Then, a simple auto-weighted robust TR decomposition algorithm was developed by leveraging an HQ technique and a scaled steepest descent method. We developed two strategies, FGMC and RStS, that exploit the special structure of the TR model to scale AWRTRD to large data. FGMC reduces the complexity of the underlying Gram matrix computation from exponential to linear, while RStS further reduces complexity by enabling the update of core tenors using a small sketch of the data. Experimental results demonstrate the superior performance of the proposed approach compared with existing TR decomposition and robust tensor completion methods.

\bibliography{ref}
\end{document}